\pdfoutput=1

\documentclass[11pt]{article}

\usepackage[]{naacl2021}

\usepackage{times}
\usepackage{latexsym}

\usepackage[T1]{fontenc}

\usepackage[utf8]{inputenc}
\usepackage{amsmath}
\usepackage{multirow}
\usepackage{caption}
\usepackage{pgfplots}
\usepackage{amssymb}
\usepackage{threeparttable}
\usepackage{array}
\usepackage{hyperref}
\usepackage{bibentry}
\usepackage{longtable}

\usepackage{microtype}
\usepackage{booktabs}
\usepackage{todonotes}
\usepackage{subcaption}

\usepackage{tikz-dependency}

\title{On Biasing Transformer Attention Towards Monotonicity}

\author{Annette Rios$^1$, Chantal Amrhein$^1$, Noëmi Aepli$^1$ \and Rico Sennrich$^{1,2}$\\
  $^1$Department of Computational Linguistics, University of Zurich\\
  $^2$School of Informatics, University of Edinburgh \\ \medskip
  \texttt{\{rios,amrhein,naepli,sennrich\}@cl.uzh.ch}}

\begin{document}
\maketitle
\begin{abstract}
Many sequence-to-sequence tasks in natural language processing are roughly monotonic in the alignment between source and target sequence, and previous work has facilitated or enforced learning of monotonic attention behavior via specialized attention functions or pretraining.
In this work, we introduce a monotonicity loss function that is compatible with standard attention mechanisms and test it on several sequence-to-sequence tasks: grapheme-to-phoneme conversion, morphological inflection, transliteration, and dialect normalization.
Experiments show that we can achieve largely monotonic behavior.
Performance is mixed, with larger gains on top of RNN baselines. 
General monotonicity does not benefit transformer multihead attention, however, we see isolated improvements when only a subset of heads is biased towards monotonic behavior.
\end{abstract}

\section{Introduction}

Many sequence-to-sequence tasks in natural language processing are roughly monotonic in the alignment between source and target sequence, and previous work has focused on learning monotonic attention behavior either through specialized attention functions \cite{aharoni-goldberg-2017-morphological, raffel2017online, wu-cotterell-2019-exact} or pretraining \cite{aji-etal-2020-neural}. 
However, it is non-trivial to port specialized attention functions to different models, and recently, \citet{Yolchuyeva2019,wu2020applying} found that a transformer model \citep{NIPS2017_3f5ee243} outperforms previous work on monotone tasks such as grapheme-to-phoneme conversion, despite having no mechanism that biases the model towards monotonicity. 

In the transformer, it is less straightforward to what extent individual encoder states, especially in deeper layers, still represent distinct source inputs after passing through several self-attention layers. Consequently, it is unclear whether enforcing monotonicity in the transformer is as beneficial as for recurrent neural networks (RNNs).

In this paper, we investigate the following research questions:

\begin{enumerate}
    \item How can we incorporate a monotonicity bias into attentional sequence-to-sequence models such as the transformer?
    \item To what extent does a transformer model benefit from such a bias?
\end{enumerate}

Specifically, we want to incorporate a monotonicity bias in a way that is agnostic of the task and model architecture, allowing for its application to different sequence-to-sequence models and tasks.
To this end, we introduce a loss function that measures and rewards monotonic behavior of the attention mechanism.\footnote{Code and scripts available at: \url{https://github.com/ZurichNLP/monotonicity_loss}}

We perform experiments and analysis on a variety of sequence-to-sequence tasks where we expect the alignment between source and target to be highly monotonic, such as grapheme-to-phoneme conversion, transliteration, morphological inflection, and dialect normalization and compare our results to previous work that successfully applied hard monotonic attention to recurrent sequence-to-sequence models for these tasks \citep{wu-etal-2018-attention, wu-cotterell-2019-exact}.

Our results show that a monotonicity bias learned through a loss function is capable of making the soft attention between source and target highly monotonic both in RNNs and the transformer. We find that this leads to a similar improvement to previous works on hard monotonic attention for RNNs, whereas for transformer models, the results are mixed: Biasing all attention heads towards monotonicity may limit the representation power of multihead attention in a way that is harmful even for monotonic sequence-to-sequence tasks. However, for some tasks, we see small improvements when limiting monotonicity to only a subset of heads.

\section{Related Work}
Attention models \cite{DBLP:journals/corr/BahdanauCB14,luong-etal-2015-effective, NIPS2017_3f5ee243} are a very powerful and flexible mechanism to learn the relationship between source and target sequences, but the flexibility might come at the cost of making the relationship harder to learn. Previous work has shown that their performance can be improved by introducing inductive biases. \citet{cohn-etal-2016-incorporating} introduce various structural alignment biases into a neural machine translation model, including a positional bias. While this bias is motivated by the fact that a given token in the source often aligns with a target token at a similar relative position, it does not explicitly encourage monotonicity. 

In contrast, \citet{raffel2017online} propose to modify the attention mechanism to learn hard monotonic alignments instead of computing soft attention over the whole source sequence. 
Several extensions have been proposed: having a pointer monotonically move over the source sequence and computing soft attention on a local window \citep{chiu2018monotonic} or from the beginning of the sequence up to the pointer \citep{arivazhagan-etal-2019-monotonic}. For tasks like simultaneous translation and automatic speech recognition, the main benefit from hard monotonic attention is that decoding becomes faster and can be done in an online setting. However, many sequence-to-sequence tasks behave roughly monotonic and biasing the attention towards monotonicity can improve performance; especially in low-resource settings. \citet{aharoni-goldberg-2017-morphological} show that hard monotonic attention works well for morphological inflection if it mimics an external alignment. 

\citet{wu-etal-2018-hard} propose a probabilistic latent-variable model for hard but non-monotonic attention which \citet{wu-cotterell-2019-exact} later extend to exact hard monotonic attention. In contrast to \citet{aharoni-goldberg-2017-morphological}, the alignment is learned jointly with the model. Their approach outperforms several other models on grapheme-to-phoneme conversion, transliteration, and morphological inflection. Monotonic attention has also improved tasks such as summarization \citep{chung2020monotonic} and morphological analysis \citep{hwang2020linear}.

Recently, the transformer architecture \citep{NIPS2017_3f5ee243} has outperformed RNNs in low-resource settings for character-level transduction tasks \citep{Yolchuyeva2019, wu2020applying} and neural machine translation \citep{araabi2020optimizing}. 
While there has been some work on extending the methods of \citet{raffel2017online, chiu2018monotonic, arivazhagan-etal-2019-monotonic} to multihead attention \citep{Ma2020Monotonic, liu2020multi}, we are not aware of any work that studied monotonicity in transformers for monotonic tasks, such as grapheme-to-phoneme conversion, transliteration, or morphological inflection. 

To this end, we propose a model-agnostic monotonicity loss that can seamlessly be integrated into RNNs as well as the transformer. Our monotonicity loss captures how monotone the soft attention behaves during training, while two hyperparameters allow us to control how much monotonicity is enforced. By encouraging monotonicity through a loss instead of a modification of the attention mechanism, our implementation still brings all the benefits of soft attention to tasks where fast, online inference is not paramount and allows us to explore various trade-offs between unconstrained and fully monotonic attention.

\section{Monotonicity Loss}

We now introduce our monotonicity loss function.
The loss function is differentiable and compatible with standard soft attention mechanisms and is thus easy to integrate into popular encoder-decoder architectures such as the transformer.
On a high level, we compare the attention distribution between decoder time steps in a pairwise fashion and measure whether the mean attended position increases for each pair.

\begin{figure*}[ht]
    \centering
    \includegraphics[width=0.8\textwidth]{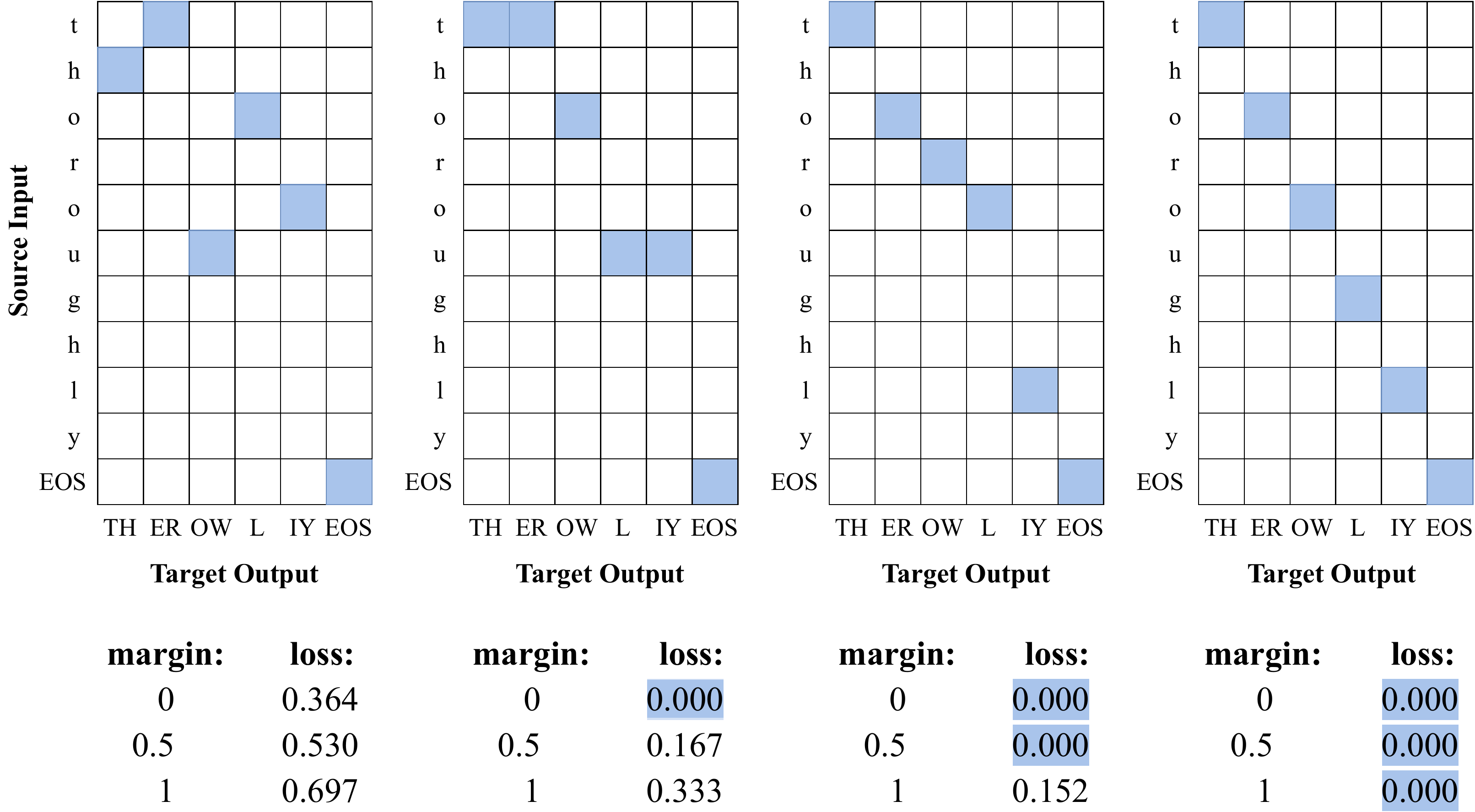}
    \caption{Average attention positions between target output characters and source input characters and the corresponding monotonicity loss for different attention distributions, and with different margins $\delta$. The average attention positions were rounded to integers for visualization purposes.}
    \label{fig:paths}
\end{figure*}

Let us denote the input sequence as $X=(x_1,...,x_{|X|})$, and the output sequence as $Y=(y_1,...,y_{|Y|})$. 
The interface between the encoder and decoder is one or several attention mechanisms.
In its general form, the attention mechanism computes some energy $e_{ij}$ between a decoder state at time step $i$ and an encoder state $j$.
While this energy function varies, with popular choices being a feedforward network \cite{DBLP:journals/corr/BahdanauCB14} or (scaled) dot-product \cite{luong-etal-2015-effective, NIPS2017_3f5ee243}, they are typically normalized to a vector of attention weights $\alpha$ using the softmax function:

\begin{equation}
\alpha_{ij} = \frac{\exp(e_{ij})}{\sum_{k=1}^{|X|}\exp(e_{ik})}
\end{equation}

These attention weights are then applied to obtain a weighted average $c_i$ of a vector of value states $V$:

\begin{equation}
c_i = \sum_{j=1}^{|x|} \alpha_{ij} \cdot v_j
\end{equation}

For our monotonicity loss, we also compute the mean attended position $\bar{a}_i$:

\begin{equation}
\bar{a}_i = \sum_{j=1}^{|x|} \alpha_{ij} \cdot j
\end{equation}

We can then define the monotonicity loss in a pairwise fashion, comparing the mean attended position at time steps $i$ and $i+1$:

\begin{equation}
L_\textrm{mono} = \sum_{i=1}^{|Y|-1} \max(\frac{\bar{a}_i - \bar{a}_{i+1} + \delta \frac{|X|}{|Y|}}{|X|}, 0)
\label{loss}
\end{equation}

$\delta$ is a hyperparameter that controls how deviations from the main diagonal are penalized.
Let us first consider the case with $\delta=0$: if \mbox{$\bar{a}_{i+1} \geq \bar{a}_{i}$} for all positions $i$, i.e.\ if the mean attended position is weakly increasing\footnote{We can swap $\bar{a}_i$ and $\bar{a}_{i+1}$ in equation \ref{loss} to bias the model towards monotonically decreasing attention.}, then the loss is 0.
Any decrease in the mean attended position will incur a cost that is proportional to the amount of decrease, relative to the source sequence length;\footnote{Making the cost relative to the source sequence length ensures that the worst-case cost per timestep is independent of source sequence length.} this allows differentiation of the loss, and will also serve as a measure of the degree of monotonicity in the analysis.

We might want to bias the model towards strictly monotonic behavior, penalizing it if $\bar{a}$ remains unchanged over several time steps.
We can achieve this by incurring a loss if $\bar{a}$ does not increase by some margin, controlled by $\delta$.
At the most extreme, with $\delta=1$, the loss is minimized if the mean attended position follows the main diagonal of the alignment matrix, increasing by $\frac{|X|}{|Y|}$ at each time step. Figure \ref{fig:paths} shows how the margin $\delta$ can influence the monotonicity loss with some examples.

In equation \ref{loss}, costs are later summed over the target sequence. In practice, we normalize the cost by the number of tokens in a batch for training stability, as is typically done for the cross-entropy loss.
If a model has multiple attention mechanisms, e.g.\ attention in multiple layers, or multihead attention, we separately compute the loss for each attention mechanism, then average the losses.
We can also just apply the loss to a subset of attention mechanisms, allowing different attention heads to learn specialized behavior \cite{voita-etal-2019-analyzing}.

\section{Experiments}

\subsection{Models and Data}

We implement the loss function in sockeye \cite{hieber-etal-2018-sockeye}, and experiment with RNN and transformer models. We list the specific baseline settings for each task in Appendix \ref{appendix:Hyperparameters}.

The monotonic loss function is controlled by a hyperparameter for the margin ($\delta$), and an additional scaling factor for the loss itself ($\lambda$).
Preliminary experiments have shown that the monotonicity loss has an undesirable interaction with attention dropout, which is commonly used in transformer models. Randomly dropping attention connections during training makes it harder to reliably avoid a decrease in the mean attended position, favoring a degenerate local optimum where attention resides constantly on the first (or last) encoder state.
To avoid this problem, we use DropHead \cite{zhou-etal-2020-scheduled} instead, which has a similar regularizing effect as attention dropout, but does not interact with the monotonicity loss. 
In addition to the standard evaluation metrics used in each task, we provide the monotonicity loss on the test set and the percentage of target tokens for which the average source attention position has increased (by some margin).

We perform experiments on three word-level and one sentence-level sequence-to-sequence tasks:

\subsubsection*{Grapheme-to-Phoneme Conversion}
For grapheme-to-phoneme conversion, we use NETtalk \citep{Sejnowski1987}\footnote{\url{https://archive.ics.uci.edu/ml/datasets/Connectionist+Bench+(Nettalk+Corpus)}} and CMUdict,\footnote{\url{https://github.com/cmusphinx/cmudict}} two datasets for English, with the same data split as \citet{wu-cotterell-2019-exact}. 
For experiments with RNN models, we follow the settings in \citet{wu-etal-2018-hard} (large configuration).\footnote{Even though we follow the settings in \citet{wu-etal-2018-hard}, our RNN models are smaller than theirs (4.5M vs.\ 8.6M parameters).} 

For experiments with transformer models, we follow the settings suggested in \citet{wu2020applying}, however, we use dropout rates of 0.3 (NETtalk) and 0.2 (CMUdict) instead of 0.1 and 0.3. Furthermore, we use a smaller feed-forward dimension for the NETtalk models (512 instead of 1024), since this a relatively small dataset ($\sim$14k samples).
 
For both RNN and transformer models, we use early stopping with phoneme error rate, as opposed to a minimum learning rate value as in \citet{wu-etal-2018-hard} and \citet{wu2020applying}. We evaluate our models with word error rate (WER) and phoneme error rate (PER).

\subsubsection*{Morphological Inflection}
For morphological inflection, we use the CoNLL-SIGMORPHON 2017 shared task dataset.\footnote{\url{https://github.com/sigmorphon/conll2017}} We choose all 51 languages from the high-resource setting where the training data for each language consists of 10,000 morphological tags + lemma and inflected form pairs (except for Bengali and Haida which have 4,243 and 6,840 pairs respectively) and from the medium-resource setting with 1,000 training examples per language. Our baselines performed very poorly on the low-resource setting with only 100 training examples and we decided to focus on the other two tasks instead. 

We preprocess the data to insert a separator token between the morphological tags and the input lemma. The monotonicity loss is then only computed on the positions to the right of the separator token's position. We follow \citet{wu2020applying} and use special positional encodings for the morphological tags in the transformer. Unlike their approach, where the position for all tags was set to 0, we set the position of the separator token to 0 and sequentially decrease the positions of the morphological tags to the left (Figure~\ref{posenc}). This serves to stabilize the positional encodings of the lemma tokens, while still accounting for the fixed order of morphological tags in the dataset. In preliminary experiments, we observed an improvement of 0.63\% in accuracy over vanilla positional encodings.

\begin{figure}
\centering
\begin{tabular}{cccccccc}
\multicolumn{8}{c}{\texttt{vanilla}} \\ \addlinespace
V & SG & 3 & PRS & <sep> & u & s & e \\
0 & 1 & 2 & 3 & 4 & 5 & 6 & 7 \\\\
\multicolumn{8}{c}{\texttt{separator-centered}} \\ \addlinespace
V & SG & 3 & PRS & <sep> & u & s & e \\
-4 & -3 & -2 & -1 & 0 & 1 & 2 & 3 \\\\
\end{tabular}
\caption{Vanilla and our proposed separator-centered positional encoding for the input ``use V;SG;3;PRS'' in the morphological inflection task.}
\label{posenc}
\end{figure}

We train models on character-level for morphological inflection following the previously recommended settings for RNNs in \citet{wu-etal-2018-hard} and for transformers in \citet{wu2020applying} (except for reducing the feed-forward dimension to 512 instead of 1024). For the high resource datasets, we use a batch size of 400, for the medium resource datasets 200. Early stopping is done in the same way as for grapheme-to-phoneme conversion. We use the official evaluation script to compute word-level accuracy (ACC) and character-level edit distance (LEV).

\subsubsection*{Transliteration}
For transliteration, we experiment on the NEWS2015 shared task data \citep{zhang-etal-2015-whitepaper} and use the same subset of 11 script pairs that \citet{wu-cotterell-2019-exact} used in their experiments: AR-EN, EN-BA, EN-HI, EN-JA, EN-KA, EN-KO, EN-PE, EN-TA, EN-TH, JN-JK, and TH-EN. Total training dataset sizes range from 6,761 source names for EN-KO up to 27,789 source names for EN-TH. For certain script pairs, multiple transliterations per source name are acceptable. We add all possible pairs to our training data, which only has a large effect on EN-AR, where there are on average 10 acceptable transliterations per source name. Since the references of the official shared task test sets were not released, we follow \citet{wu-cotterell-2019-exact} and use the development set as our test set. We randomly sample 1,000 names from the training sets as our development sets for script pairs with more than 20,000 training examples and 100 for script pairs with fewer training examples.

Again, we follow \citet{wu-etal-2018-hard} for hyperparameters in RNNs and \citet{wu2020applying} in transformers (smaller feed-forward dimensions of 512). We early stop training as for grapheme-to-phoneme conversion.  We evaluate our models following \citet{zhang-etal-2015-whitepaper} and compute word-level accuracy (ACC) and character-level mean F-score (MFS). The formula for MFS is in Appendix \ref{appendix:mfs}.

\subsubsection*{Dialect Normalization}
\begin{figure}
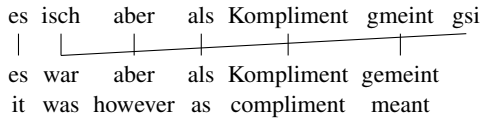

\centering
\begin{dependency}[theme = simple]
   \begin{deptext}[column sep=0pt, font=\small]
      es \& isch \& aber \& als \& Kompliment \& gmeint \& gsi\\
      \\
      es \& war \& aber \& als \& Kompliment \& gemeint\\
      it \& was \& however \& as \& compliment \& meant\\
   \end{deptext}
    \begin{scope}
    \draw (\wordref{1}{1}.south)--(\wordref{3}{1}.north);
    \draw (\wordref{1}{2}.south)--(\wordref{3}{2}.north);
    \draw (\wordref{1}{3}.south)--(\wordref{3}{3}.north);
    \draw (\wordref{1}{4}.south)--(\wordref{3}{4}.north);
    \draw (\wordref{1}{5}.south)--(\wordref{3}{5}.north);
    \draw (\wordref{1}{6}.south)--(\wordref{3}{6}.north);
    \draw (\wordref{1}{7}.south)--(\wordref{3}{2}.north);
  \end{scope}
\end{dependency}
\caption{Swiss-German to German dialect normalization example with verb reordering.}
\label{reorder}
\end{figure}

For this work, we consider dialect normalization as a machine translation task from dialect to standard. We work with the dataset described in \citet{zora159152}, which consists of 26,015 crowd-sourced German translations of 6,197 original Swiss German sentences. We use three documents (10\%) as test sets and randomly split the rest in development and training set (10\% and 80\% respectively).
The alignment between Swiss German and the German translations is highly monotonic, but there are occasional word order differences, as illustrated in Figure~\ref{reorder}.

The models are trained on subwords obtained via BPE \citep{sennrich-etal-2016-neural}, created with subword-nmt computing 2000 merges.
We treat this as a low-resource machine translation task, and thus follow hyperparameters by \citet{sennrich-zhang-2019-revisiting} for the RNN models, while the transformer models are trained according to \citet{araabi2020optimizing}. We evaluate our models with BLEU \citep{papineni-etal-2002-bleu}.\footnote{SacreBLEU \citep{post-2018-call}: BLEU+case.mixed+numrefs.1\\+smooth.exp+tok.13a+version.1.3.6}

\begin{table*}
    \centering
    \small
    \setlength{\tabcolsep}{0.4em}
    
         \begin{tabular}{rrrrrrrrr}
        & \multicolumn{8}{c}{Grapheme-to-Phoneme Conversion}  \\
          \cmidrule(lr){2-9}  
         & &  & \multicolumn{2}{c}{$\delta=0.0$} & \multicolumn{2}{c}{$\delta=0.5$} & \multicolumn{2}{c}{$\delta=1.0$}  \\
         & WER $\downarrow$& PER $\downarrow$ &  \small{$\%mono$} &\small{$L_{MONO}$}  & \small{$\%mono$}& \small{$L_{MONO}$}  & \small{$\%mono$} & \small{$L_{MONO}$} \\
           \cmidrule(lr){2-2} \cmidrule(lr){3-3} \cmidrule(lr){4-4} \cmidrule(lr){5-5} \cmidrule(lr){6-6} \cmidrule(lr){7-7} \cmidrule(lr){8-8} \cmidrule(lr){9-9} \addlinespace 
            \textbf{RNN}   \\
         \citet{wu-etal-2018-hard} & 28.20\phantom{$\pm$0.00} & 6.8\phantom{0$\pm$0.00} & \\
         \citet{wu-cotterell-2019-exact} & 28.20\phantom{$\pm$0.00} & 6.9\phantom{0$\pm$0.00} &  \\ 
         \addlinespace
          baseline ($\lambda=0$) & 28.76$\pm$0.73 & 7.16$\pm$0.16  & 84.7\% & 2.91e-04 &  84.0\%\ & 3.94e-03  & 26.8\% & 1.03e-01  \\  \addlinespace
        $\lambda=0.1$, $\delta=0.0$ & 28.88$\pm$0.32 & 7.16$\pm$0.03  & 84.8\% & 1.24e-04 \\ 
        $\lambda=0.1$, $\delta=0.5$  & \textbf{28.55}$\pm$0.18 & \textbf{7.13}$\pm$0.09 & & & 84.3\% & 1.74e-03 \\
        $\lambda=0.1$, $\delta=1.0$  & 29.02$\pm$0.55 & 7.32$\pm$0.19 & & & &  & 44.5\% & 4.05e-02\\ \addlinespace 
      
          \textbf{Transformer} \\
           \citet{wu2020applying} &  27.63\phantom{$\pm$0.00} & 6.9\phantom{0$\pm$0.00} & \\ \addlinespace
          baseline  & 27.79$\pm$0.24 & 7.00$\pm$0.09 & 77.0\%  & 7.26e-02\\ \addlinespace
         $\lambda=0.1$, $\delta=0.0$, $h$ = all: & 27.99$\pm$0.60 & 7.11$\pm$0.18 &  84.6\% & 5.12e-05\\
       
      \end{tabular}
         \caption{Results for grapheme-to-phoneme, monotonicity loss for transformer on all layers and heads. Average over three runs with independent seeds with $\lambda=0.1$. Our best models are marked in bold.}
          \label{tab:g2p}
\end{table*}

\begin{table*}
    \centering
    \small
    \setlength{\tabcolsep}{0.4em}
        \begin{tabular}{rcccccccc}
         & \multicolumn{4}{c}{Morph. Infl. High Resource} & \multicolumn{4}{c}{Morph. Infl. Medium Resource}  \\
          \cmidrule(lr){2-5}  \cmidrule(lr){6-9} 
         & ACC $\uparrow$ & LEV $\downarrow$ & $\%mono$  & $L_{MONO}$ & ACC $\uparrow$ & LEV $\downarrow$ & $\%mono$  & $L_{MONO}$\\ \cmidrule(lr){2-2} \cmidrule(lr){3-3} \cmidrule(lr){4-4} \cmidrule(lr){5-5} \cmidrule(lr){6-6} \cmidrule(lr){7-7} \cmidrule(lr){8-8} \cmidrule(lr){9-9} \addlinespace
         \textbf{RNN} \\
         \citet{wu-etal-2018-hard} & 93.60\phantom{$\pm$0.00} & 0.128\phantom{$\pm$0.000}  & - & -\\
         \citet{wu-cotterell-2019-exact} & 94.81\phantom{$\pm$0.00} & 0.123\phantom{$\pm$0.000}  & - & -\\ 
         baseline ($\lambda$ = 0) & \textbf{94.97}$\pm$0.06 & \textbf{0.098}$\pm$0.002 & 65.5\% &  1.17 & \textbf{78.15}$\pm$0.24 & \textbf{0.441}$\pm$0.005 & 64.6\% & 1.16 \\
         $\lambda$ = 0.1, $\delta$ = 0.0 & 94.63$\pm$0.01 &  0.105$\pm$0.002 & 83.5\% & 3.78e-4 & 74.11$\pm$0.35 & 0.560$\pm$0.009 & 84.3\% & 1.63e-3\\
         \addlinespace
         \textbf{Transformer} \\
         \citet{wu2020applying} & 95.59\phantom{$\pm$0.00} & 0.088\phantom{$\pm$0.000} & - & -\\
         baseline ($\lambda$ = 0) & \textbf{95.05}$\pm$0.03 & \textbf{0.097}$\pm$0.001 & 58.1\% & 1.34 &  \textbf{81.33}$\pm$0.02 & \textbf{0.378}$\pm$0.001 & 58.1\% & 1.35 \\
         $\lambda$ = 0.1, $\delta$ = 0.1, $h$ = all & 94.98$\pm$0.07 & 0.099$\pm$0.002 & 87.5\% & 4.49e-4 & 81.02$\pm$0.17 & 0.383$\pm$0.001 & 85.7\% & 1.45e-3
         \end{tabular}

         \caption{Results for morphological inflection, monotonicity loss for transformer on all layers and heads. Average over three runs with independent seeds with $\lambda=0.1$. Our best models are marked in bold.}
         \label{tab:morph}

\end{table*}

\begin{table*}
    \centering
    \small
        \setlength{\tabcolsep}{0.4em}
        \begin{tabular}{rccccccc}
         & \multicolumn{4}{c}{Transliteration} &  \multicolumn{3}{c}{Dialect Normalization} \\
          \cmidrule(lr){2-5}   \cmidrule(lr){6-8} 
         & ACC $\uparrow$ & MFS $\uparrow$ & $\%mono$ & $L_{MONO}$ & BLEU $\uparrow$ & $\%mono$ &  $L_{MONO}$\\ \cmidrule(lr){2-2} \cmidrule(lr){3-3} \cmidrule(lr){4-4} \cmidrule(lr){5-5}  \cmidrule(lr){6-6} \cmidrule(lr){7-7} \cmidrule(lr){8-8} \addlinespace
         \textbf{RNN} \\
         \citet{wu-etal-2018-hard} & 41.10\phantom{$\pm$0.00} & 89.40\phantom{$\pm$0.00} & - & -\\
         \citet{wu-cotterell-2019-exact} & 41.20\phantom{$\pm$0.00} & 89.50\phantom{$\pm$0.00} & - & -\\ 
         baseline ($\lambda$ = 0) & 39.53$\pm$0.56 & 89.06$\pm$0.06 & 74.4\% & 0.06 &  \textbf{33.41}$\pm$0.39 & 83.1\% & 0.42 \\
         $\lambda$ = 0.1, $\delta$ = 0.0 & \textbf{40.03}$\pm$0.39 & \textbf{89.18}$\pm$0.04 & 81.7\% & 1.4e-3 & 33.29$\pm$0.23 & 90.2\% & 0.10\\
         \addlinespace
         \textbf{Transformer} \\
         \citet{wu2020applying} & 43.39\phantom{$\pm$0.00} & 89.70\phantom{$\pm$0.00} & - & -\\
         baseline ($\lambda$ = 0) & \textbf{42.08}$\pm$0.55 & \textbf{89.63}$\pm$0.04 & 69.1\%  & 0.12 & \textbf{32.83}$\pm$0.20 &  71.7\%  & 1.23\\
         $\lambda$ = 0.1, $\delta$ = 0.0, $h$ = all& 41.32$\pm$0.53 & 89.47$\pm$0.08 & 82.2\% & 7.1e-4 & 32.17$\pm$0.78 & 91.1\% & 0.05 \\
         \end{tabular}
     \caption{Results for transliteration and dialect normalization, all experiments with $\delta$ = 0. Monotonicity Loss for transformer on all layers and heads. Average over three runs with independent seeds with $\lambda=0.1$. Our best models are marked in bold.}
         \label{tab:transliteration}
         
\end{table*}

\subsection{Results}
In addition to task-specific evaluation metrics, we use the loss function to score the monotonicity of the attention on the test set for all models (reported as $L_{MONO}$). 
Furthermore, we report the percentage of decoding states for which the average source attention position $\bar{a}$ increases by at least $\delta \frac{|X|}{|Y|}$ as $\%mono$. In other words, this is the percentage of states for which the pairwise loss is 0.

\subsubsection*{Grapheme-to-Phoneme Conversion}
We test different settings on the grapheme-to-phoneme task, see Table \ref{tab:g2p} for results with RNNs (top) and transformers (bottom).
We find that models trained with the additional loss have more monotonic attention than the baselines (see $\%mono$ and $L_{MONO}$).
We observe large differences both in terms of WER and PER across multiple runs for the baseline, especially for the small data set.\footnote{Standard deviation on NETtalk with RNN is >1.2 WER and >0.27 PER across baseline runs.} We therefore report the average result of three runs with standard deviations for each model.

Attention in the RNN baselines is already quite monotonic, but we observe small improvements with $\delta$ = 0.5.
For transformer models, on the other hand, $\delta>0$ seems to harm the performance, therefore we only report results with $\delta = 0$.
In general, multihead attention in the transformer does not seem to benefit much from enforced monotonicity.

\subsubsection*{Morphological Inflection}
For morphological inflection, we show the average results over all 51 languages in Table \ref{tab:morph}. Our RNN baseline is slightly better than previous work, whereas our transformer baseline performs slightly worse. We notice that the transformer models trained with $\delta=0$ on the morphological inflection tasks result in the model always attending to the same source position at every decoding state. We therefore set $\delta$ to 0.1 for transformer models trained on this task. For the remaining tasks, we report results with $\delta$ set to 0 and $\lambda$ always set to 0.1 so as not to overfit hyperparameters on each task.

The baseline monotonicity loss for this task is higher than for grapheme-to-phoneme conversion but training with the monotonicity loss can drastically increase the monotonicity of the attention mechanisms. This can be seen both in the lower monotonicity score and the higher percentage of decoding states where the average source attention position increases from the previous state. In terms of performance, we do not see an improvement over the baselines. 

\subsubsection*{Transliteration}

Our results for transliteration are shown in Table \ref{tab:transliteration} (average over all 11 datasets). Again, we can see that the monotonicity loss effectively biases the attention towards a more monotonic behavior, decreasing the monotonicity score and increasing the percentage of decoding states where the average source attention position increases. In terms of performance, there is a small gain for RNNs both in word-level accuracy and character-level mean F-score. Training with the monotonicity loss does not improve the performance of the transformer compared to the baseline. 

\subsubsection*{Dialect Normalization}

Since dialect normalization is our only sentence-level sequence-to-sequence task, it is interesting to see how the monotonicity loss works on longer sequences where more reordering is possible compared to the previous tasks. The less monotonic nature of this task is reflected in the fact that neither of our models trained towards monotonicity outperforms the non-monotonic baselines, see Table \ref{tab:transliteration}. Dialect normalization is also the only task where the transformer does not outperform the RNN models. 

\section{Analysis}
Overall, our results show that the proposed monotonicity loss succeeds in making attention more monotonic, but effects on quality are more positive for RNNs than for transformers.
We now analyze the proposed loss function in more detail.

\subsection*{Monotonicity Over Time}
\begin{figure}
    \centering
    \includegraphics[width=0.35\textwidth]{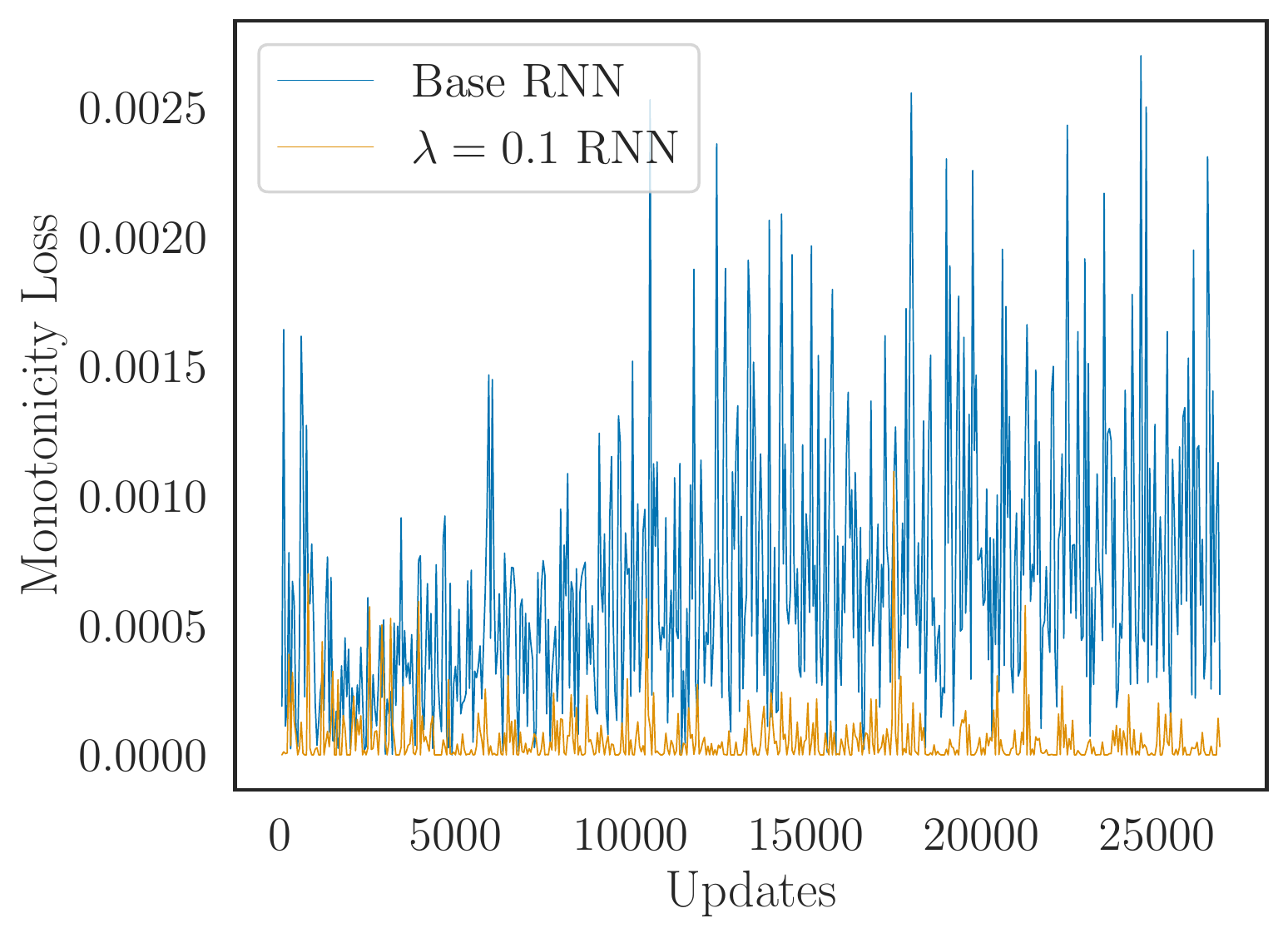}
    \includegraphics[width=0.35\textwidth]{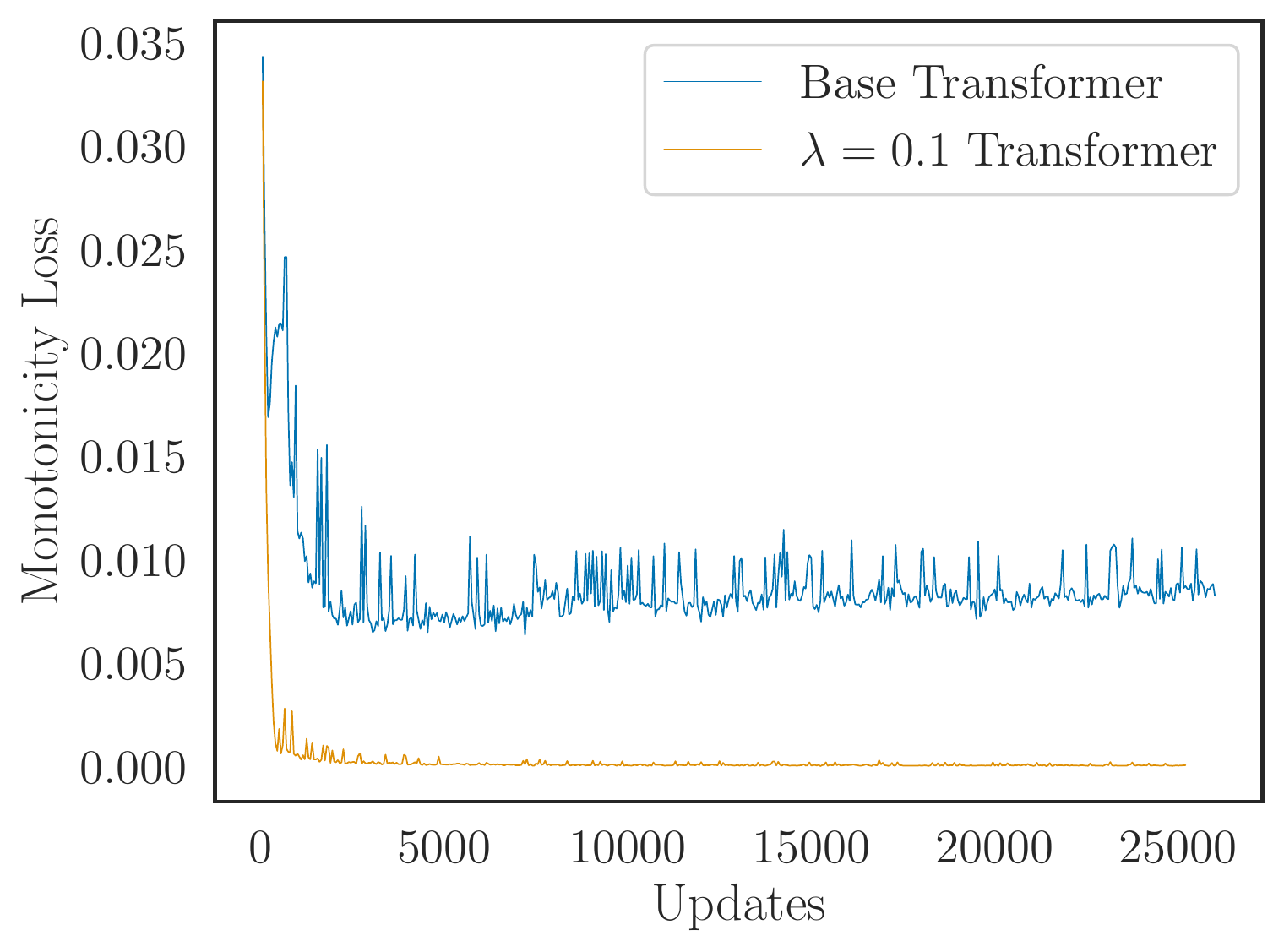}
    \caption{Monotonicity score during training on the EN-JA transliteration dataset with $\delta=0$. Upper plot: RNN, lower plot: transformer (all heads).}
    \label{fig:train_loss}
\end{figure}

First, we plot the monotonicity score during training and compare how fast it decreases over time. We find that the monotonicity score decreases very fast for the models trained with our loss function and then stays rather constant. The baseline models show various behaviors: for some datasets and models, the score decreases over training time - suggesting that the model does learn to attend more monotonically even without the loss. For other data sets, the score is initially lower and increases over training time, and, for some, the score stays more or less constant. What all baselines have in common, is that the monotonicity score oscillates much more than when trained with the monotonicity loss. Figure \ref{fig:train_loss} shows an example plot for the EN-JA transliteration dataset.

\subsection*{Varying Monotonicity}
\begin{figure}
    \centering
    \includegraphics[width=0.35\textwidth]{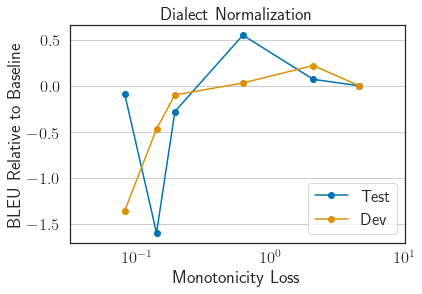}
    \caption{Relative BLEU scores as a function of the monotonicity loss for dialect normalization with transformer (all heads). Different data points obtained by varying $\lambda$. ($\lambda \in \{0.3,0.2,0.1,0.01,0.001,0\}$).}
    \label{fig:bleu_vs_loss}
\end{figure}

We can vary how much we constrain attention to be monotonic by varying
the weight of the monotonicity loss function ($\lambda$). We analyze how this influences the performance on dialect normalization. Figure~\ref{fig:bleu_vs_loss} shows that non-monotonic behavior (as defined by the monotonicity loss) can be reduced by a factor of 10-20 with stable or even slightly improving performance.
However, BLEU drops drastically for large $\lambda$. This highlights the advantage of our loss function over hard monotonic attention. Through $\lambda$ we can regulate the degree of monotonicity in the attention mechanism, which can be beneficial for tasks where hard monotonic attention would be too strict.

\begin{table*}[t]
    \centering
    \small
    \setlength{\tabcolsep}{0.4em}
        \begin{tabular}{rcccccc}
          & \multicolumn{2}{c}{Performance} &  \multicolumn{2}{c}{heads with $L_{MONO}$} & \multicolumn{2}{c}{heads without $L_{MONO}$}\\ \cmidrule(lr){2-3}  \cmidrule(lr){4-5} 
          \cmidrule(lr){6-7} \addlinespace
         \textbf{G2P} & WER $\downarrow$ & PER $\downarrow$ & $\%mono$ & $L_{MONO}$ & $\%mono$ & $L_{MONO}$\\ 
         \cmidrule(lr){2-2} \cmidrule(lr){3-3} \cmidrule(lr){4-4} \cmidrule(lr){5-5} \cmidrule(lr){6-6} \cmidrule(lr){7-7}  \addlinespace
         baseline  & 27.79$\pm$0.24 & 7.00$\pm$0.09 & & & 77.0\%  & 7.26e-02\\
         $\lambda=0.1$, $\delta=0.0$, $h$ = all: & 27.99$\pm$0.60 & 7.11$\pm$0.18 & 84.6\% & 5.12e-05\\
         $\lambda=0.1$, $\delta=0.0$, $h$ = 1\phantom{ll}: & \textbf{27.70}$\pm$0.37 & \textbf{6.96}$\pm$0.07 & 84.9\%  & 2.49e-05 & 75.1\% & 8.26e-02\\\addlinespace\addlinespace
         \textbf{Morph. Infl. High} & ACC $\uparrow$ & LEV $\downarrow$ \\ 
         \cmidrule(lr){2-2} \cmidrule(lr){3-3}  \addlinespace
         baseline  & \textbf{95.05}$\pm$0.03 & \textbf{0.097}$\pm$0.001 & & & 58.1\% & 1.34  \\
         $\lambda=0.1$, $\delta=0.1$, $h$ = all: & 94.98$\pm$0.07 & 0.099$\pm$0.002 & 87.5\% & 4.49e-4\\
         $\lambda=0.1$, $\delta=0.1$, $h$ = 1\phantom{ll}: & 95.00$\pm$0.03 & 0.098$\pm$0.000  & 89.3\% & 6.49e-5 & 59.6\% & 1.35 \\\addlinespace\addlinespace
         \textbf{Morph. Infl. Medium} & ACC $\uparrow$ & LEV $\downarrow$ \\ 
         \cmidrule(lr){2-2} \cmidrule(lr){3-3}  \addlinespace
         baseline  &  81.33$\pm$0.02 & 0.378$\pm$0.001 & & & 58.1\% & 1.35\\
         $\lambda=0.1$, $\delta=0.1$, $h$ = all: & 81.02$\pm$0.17 & 0.383$\pm$0.001 & 85.7\% & 1.45e-3 \\
         $\lambda=0.1$, $\delta=0.1$, $h$ = 1\phantom{ll}: &  \textbf{81.67}$\pm$0.13 & \textbf{0.366}$\pm$0.003 & 88.6\% & 4.28e-4 & 59.0\% & 1.37  \\\addlinespace\addlinespace
         \textbf{Transliteration} & ACC $\uparrow$ & MFS $\uparrow$ \\ 
         \cmidrule(lr){2-2} \cmidrule(lr){3-3}  \addlinespace
         baseline  &  \textbf{42.08}$\pm$0.55 & \textbf{89.63}$\pm$0.04 & & & 69.1\%  & 0.12  \\
         $\lambda=0.1$, $\delta=0.0$, $h$ = all: & 41.32$\pm$0.53 & 89.47$\pm$0.08 & 82.2\% & 7.1e-4 \\
         $\lambda=0.1$, $\delta=0.0$, $h$ = 1\phantom{ll}: &  41.71$\pm$0.37 & 89.61$\pm$0.06 & 80.4\% & 1.12e-4 & 69.6\% & 0.11 \\\addlinespace\addlinespace
         \textbf{Dialect Normalization} & \multicolumn{2}{c}{BLEU $\uparrow$} \\ 
         \cmidrule(lr){2-3}  \addlinespace
         baseline  & \multicolumn{2}{c}{\textbf{32.83}$\pm$0.20} & & & 71.7\% & 1.23 \\
         $\lambda=0.1$, $\delta=0.0$, $h$ = all: & \multicolumn{2}{c}{32.17$\pm$0.78} & 91.1\% & 0.05 \\
         $\lambda=0.1$, $\delta=0.0$, $h$ = 1\phantom{ll}: &\multicolumn{2}{c}{31.55$\pm$0.71} & 77.9\%  & 0.01 & 70.5\% & 1.64 \\\addlinespace
         \end{tabular}
         \caption{Transformer results for all tasks with monotonicity on all heads vs.\ only on one head. Monotonicity loss is computed on all layers. Average over three runs with independent seeds. Our best models are marked in bold.}
         \label{tab:one_head}
         
\end{table*}

\subsection*{Monotonicity Loss on Single Heads}
Since we calculate the loss on each attention component separately, we can also limit its application to specific layers and heads (in the case of multihead attention). We test how restricting the monotonic behavior to only one head per layer influences the performance of the transformer on our chosen tasks. Results are presented in Table \ref{tab:one_head}. We find that monotonicity on only one head generally improves performance compared to on all heads, except for dialect normalization. For grapheme-to-phoneme conversion and morphological inflection in the medium resource setting, we even see performance gains over the baseline. 

Our results support the belief that the flexibility of multihead attention is key to the success of the transformer. If applied to all heads, the monotonicity loss reduces variability in the attention distribution of the different heads, i.e. with high $\lambda$, all heads attend to the same source position. We suspect that this severely limits the capacity of transformer models and explains why rewarding monotonicity on only one head is beneficial.

These findings are also important in the context of the work by \citet{voita-etal-2019-analyzing} who find that attention heads tend to learn specialized functions. Having one monotonic attention head could be a complementary way to encourage more diversity amongst heads, next to disagreement regularization \cite{li-etal-2018-multi-head}. Indeed, we observe that for grapheme-to-phoneme conversion and dialect normalization the remaining heads trained without the monotonicity loss tend to become less monotonic.

\subsection*{Attention Maps}
\begin{figure}
    \centering
    \includegraphics[width=0.48\textwidth]{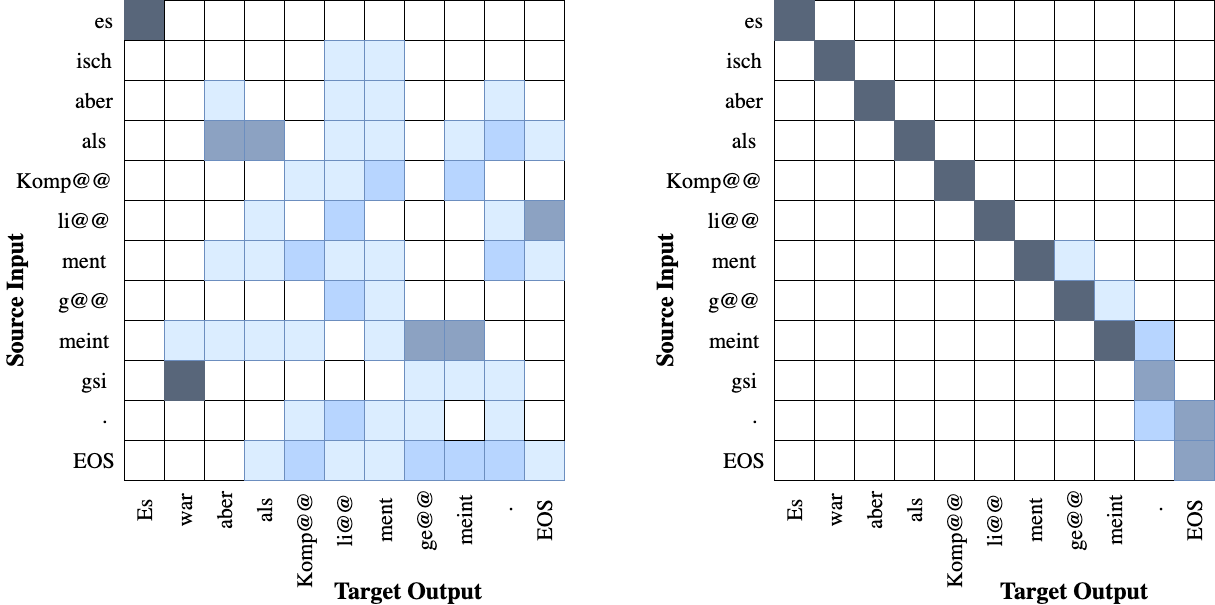}
    \includegraphics[width=0.48\textwidth]{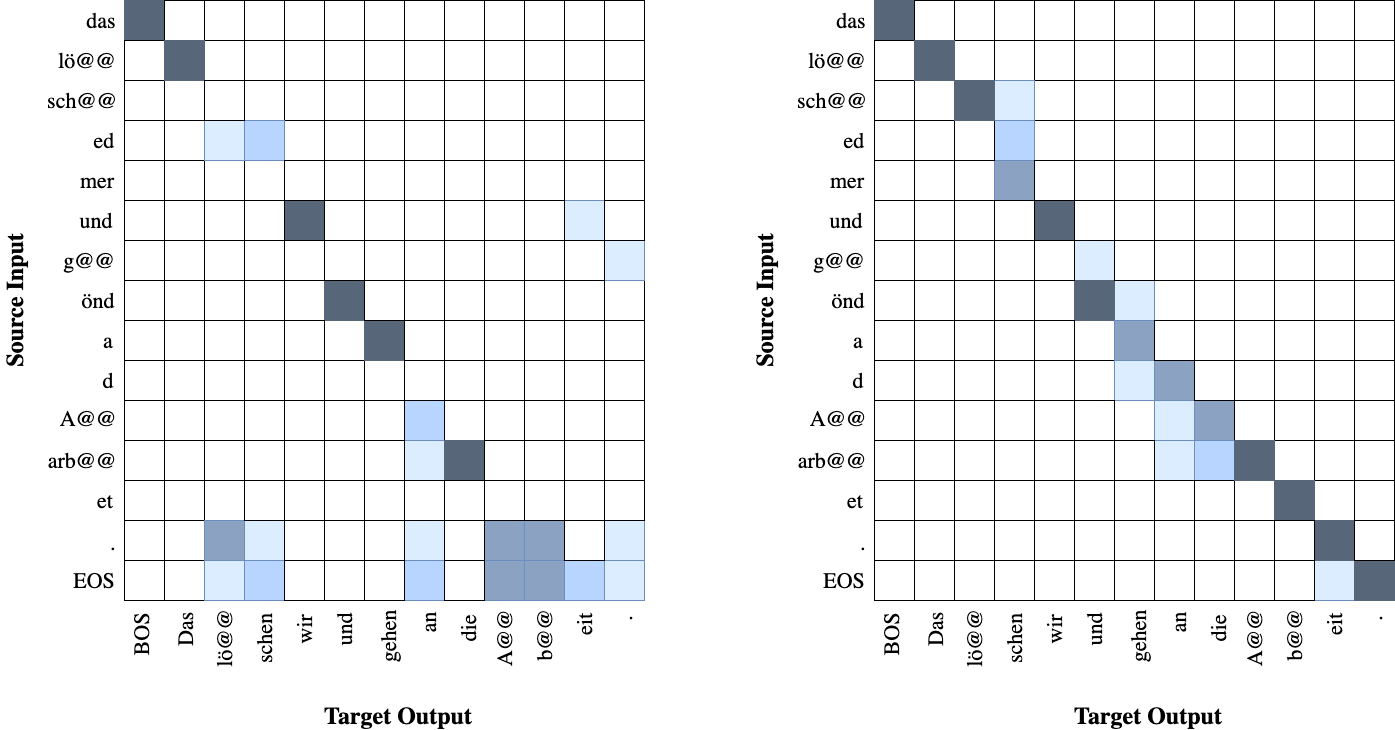}
    \caption{Transformer attention maps for the sentence shown in Figure \ref{reorder}; \textit{``but it was meant as  compliment"} and \textit{``we delete this and get to work"}. Left: baseline ($\lambda$=0), right: with monotonicity loss on all heads ($\lambda$=0.1).}
    \label{fig:attention_map_DN}
\end{figure}

Attention maps are particularly interesting for dialect normalization where 1) the transformer baseline has one of the highest monotonicity losses of all our models and 2) reordering of source and target tokens is possible. Figure \ref{fig:attention_map_DN} shows the attention maps for our baseline transformer and the corresponding model trained with the monotonicity loss. 
The bottom sentence is an example where the alignment between the source and the target is monotonic. Here, the baseline does show tentative monotonic behavior but with the monotonicity loss, the attention follows the main diagonal much more closely. 
The sentence on the top, on the other hand, contains a non-monotonic alignment. For a correct alignment of the past tense of ``to be'', the model needs to peek at the very last token before the full stop. This is reflected in the baseline attention map where the attention at the second decoding step is highest on the third-to-last source position. However, for our model trained with the monotonicity loss, the attention follows the main diagonal and fails to mirror the correct alignment. Occasional reorderings like this may explain why the monotonicity loss did not work well for this task despite it being largely monotonic.

\section{Conclusion}
We propose a model-agnostic loss function that measures and rewards monotonicity and can easily be integrated into various attention mechanisms. To achieve this, we track how monotonically the average position of the attention shifts over the source sequence across time steps. We show that this loss function can be seamlessly integrated into RNNs as well as transformers. Models trained with our monotonicity loss learn largely monotonic behavior without any specific changes to the attention mechanism.  While we see some performance gains in RNNs, our results show that biasing all attention heads in transformers towards monotonic behavior is undesirable.
However, a bias towards monotonicity may be helpful if applied to only a subset of heads. 

For the future, we are interested in more sophisticated schedules for the monotonicity loss, possibly reducing $\lambda$ over the course of training. This would help to learn monotonic behavior in the early training stages but gives the model more flexibility to deviate from such an attention pattern if needed. In this context, our loss function could also be used as an additional pretraining objective for transfer to very low-resource tasks. We would also like to test our loss function on tasks where the alignment may be harder to learn, for example in multimodal models or for long sequences. Finally, using our loss function as a way to measure monotonicity could be an interesting tool for interpretability research.

\section*{Acknowledgments}

We thank the anonymous reviewers for their feedback.
This project has received funding from the Swiss National Science Foundation (project nos.\ 176727 and 191934).

\clearpage
\bibliography{anthology,custom}
\bibliographystyle{acl_natbib}

\clearpage
\appendix
\onecolumn
\section{Appendix}
\subsection{Character-level Mean F-score (MFS)}\label{appendix:mfs}
\vspace{1cm}
\begin{equation*}
    LCS(c_i,r_i) = \frac{1}{2} (\lvert c_i \rvert + \lvert r_i \rvert - ED(c_i,r_i))
\end{equation*}

\begin{equation*}
    R_i = \frac{LCS(c_i,r_i)}{\lvert r_i \rvert}
\end{equation*}

\begin{equation*}
    P_i = \frac{LCS(c_i,r_i)}{\lvert c_i \rvert}
\end{equation*}

\begin{equation*}
    F_i = \frac{2 * R_i * P_i}{R_i + P_i}
\end{equation*}

\begin{equation*}
    MFS = \frac{1}{N} \sum_{i=1}^{N} F_i
\end{equation*}\\
 
 \noindent Where $c_i$ is the i-th candidate and $r_i$ is the corresponding reference transliteration with the smallest edit distance (ED).  
  
\newpage

\subsection{Hyperparameters}\label{appendix:Hyperparameters}
\nopagebreak

      \begin{longtable}{lcccc}
        \toprule
         \multicolumn{5}{c}{Training Hyperparameters Transformer} \\ \midrule
          & G2P & MI & TR & DN \\ 
         training settings: \\ \cmidrule{1-1}
         batch type & \multicolumn{3}{c}{sentence} & word \\
         batch size  & 400 & 400/200 & 400 & 4096 \\
         max-seq-len & 20:20  & 85:85 & 85:85  & 200:200 \\
         word-min-count &  \multicolumn{4}{c}{1:1}\\
         seed & \multicolumn{4}{c}{1, 2, 3} \\ \midrule \addlinespace
         
         model settings: \\ \cmidrule{1-1}
         encoder &  \multicolumn{4}{c}{transformer} \\
         decoder &  \multicolumn{4}{c}{transformer} \\
         transformer-positional-embedding-type &  \multicolumn{4}{c}{fixed} \\
         transformer-preprocess & \multicolumn{4}{c}{n} \\
         transformer-postprocess & \multicolumn{4}{c}{dr} \\
         num-layers & \multicolumn{3}{c}{4:4} & 5:5 \\
         transformer-model-size & \multicolumn{3}{c}{256} & 512 \\
         transformer-attention-heads & \multicolumn{3}{c}{4} & 2\\
         num-embed & \multicolumn{3}{c}{256:256} & 512:512\\
         weight-tying-type & \multicolumn{3}{c}{trg\_softmax} & src\_trg \\ 
         transformer-feed-forward-num-hidden & 512/1024  & 512 & 512 & 512 \\ \midrule \addlinespace
         
         optimization settings: \\ \cmidrule{1-1}
         optimizer & \multicolumn{4}{c}{adam} \\
         optimizer-params &  \multicolumn{4}{c}{beta2:0.98} \\
         checkpoint interval & \multicolumn{4}{c}{400} \\
         max-num-checkpoint-not-improved & \multicolumn{4}{c}{10} \\ 
         gradient-clipping-threshold &  \multicolumn{4}{c}{none}\\
         learning-rate-scheduler-type &\multicolumn{3}{c}{fixed-rate-inv-sqrt-t} & plateau-reduce\\
         optimized-metric & \multicolumn{3}{c}{PER} & bleu  \\
         label-smoothing & \multicolumn{3}{c}{0.1} & 0.6\\ 
         initial-learning-rate & \multicolumn{3}{c}{0.001} & 0.0001\\
         learning-rate-warmup & \multicolumn{3}{c}{4000} & 0\\ \midrule \addlinespace
      
         initialization settings: \\ \cmidrule{1-1}
         weight-init &  \multicolumn{4}{c}{xavier} \\
         weight-init-scale &  \multicolumn{4}{c}{3.0} \\
         weight-init-xavier-factor-type & \multicolumn{4}{c}{avg} \\ \midrule \addlinespace
        
         dropout settings:  \\ \cmidrule{1-1}
         transformer-dropout-attention &  \multicolumn{4}{c}{0}\\
         embed-dropout &  0.3/0.2 & 0.3 & 0.3 & 0.1\\
         transformer-drophead-attention &  0.3/0.2 & 0.3 & 0.3 & 0.0/0.1\\
         transformer-dropout-act & 0.3/0.2 & 0.3 & 0.3 & 0.3\\
         transformer-dropout-prepost & 0.3/0.2 & 0.3 & 0.3 & 0.3\\        
         \bottomrule \\
          \caption{Sockeye hyperparameters for transformer models (values with ':' = encoder:decoder)} 
        \end{longtable}
 \pagebreak
 
  \begin{longtable}{lcccc}
        \toprule
         \multicolumn{5}{c}{Training Hyperparameters RNN} \\ \midrule
          & G2P & MI & TR & DN \\ 
          \cmidrule(lr){2-2} \cmidrule(lr){3-3} \cmidrule(lr){4-4} \cmidrule(lr){5-5}
         training settings: \\ \cmidrule{1-1}
         batch type &   \multicolumn{3}{c}{sentence} &  word \\
         batch size &  20 & 20  & 50  & 1000\\
         max-seq-len & 20:20 & 85:85 & 85:85   & 200:200\\
         word-min-count &   \multicolumn{4}{c}{1:1}  \\
         seeds & \multicolumn{4}{c}{1, 2, 3} \\ \midrule \addlinespace
         
         model settings: \\ \cmidrule{1-1}
         encoder &    \multicolumn{4}{c}{rnn}  \\
         decoder &     \multicolumn{4}{c}{rnn}  \\
         rnn-cell-type &  \multicolumn{4}{c}{lstm} \\
         num-layers &  \multicolumn{3}{c}{2:1}& 1:1\\
         num-embed & \multicolumn{3}{c}{200:200} & 512:512\\
         rnn-num-hidden  & \multicolumn{3}{c}{400} & 1024\\  \midrule \addlinespace
         
         optimization settings: \\ \cmidrule{1-1}
         learning-rate-scheduler-type &  \multicolumn{4}{c}{plateau-reduce} \\
         learning-rate-warmup &  \multicolumn{4}{c}{0} \\ 
         optimizer & \multicolumn{4}{c}{adam} \\
         optimized-metric & \multicolumn{3}{c}{PER}  & bleu\\
         checkpoint interval &  \multicolumn{3}{c}{4000} & 400\\
         max-num-checkpoint-not-improved & \multicolumn{3}{c}{7} & 10\\ 
         label-smoothing & \multicolumn{3}{c}{0.0} & 0.2\\ 
         gradient-clipping-threshold & \multicolumn{3}{c}{5} & --\\
         initial-learning-rate & \multicolumn{3}{c}{0.001} & 0.0005\\
         learning-rate-reduce-num-not-improved &  \multicolumn{3}{c}{1}  & 8\\
         learning-rate-reduce-factor & \multicolumn{3}{c}{0.5}  & 0.7\\\midrule \addlinespace
      
         initialization settings: \\ \cmidrule{1-1}
         weight-init &  \multicolumn{4}{c}{xavier} \\
         weight-init-scale & \multicolumn{4}{c}{3.0} \\
         weight-init-xavier-factor-type &\multicolumn{4}{c}{avg} \\ \midrule \addlinespace
        
         dropout settings:  \\ \cmidrule{1-1}
         embed-dropout &  \multicolumn{3}{c}{0.4} & 0.5\\
        rnn-decoder-hidden-dropout & \multicolumn{3}{c}{0.4} & 0.5\\
         \bottomrule \\
          \caption{Sockeye hyperparameters for RNN models (values with ':' = encoder:decoder)} 
        \end{longtable}

        \pagebreak
        \subsection{Model Size}\label{appendix:size}
\nopagebreak
        \begin{table}[h]
            \centering
            \begin{tabular}{lr}
            \toprule
                 & RNN Models \\ \cmidrule{2-2}
               G2P  & 4.5M \\
               MI & 4.5M\\
               TR & 4.5M\\
               DN & 25.1M\\   \addlinespace
               & Transformer Models \\  \cmidrule{2-2}
               G2P   \\ 
               \hspace{3mm}CMUdict (ff = 1024): & 7.3M \\
               \hspace{3mm}NETtalk (ff = 512):&  5.3M \\
               \addlinespace
               MI & 5.3M\\
               TR & 5.3M\\
               DN & 23.2M\\ \addlinespace
               \bottomrule
            \end{tabular}
            \caption{Approximate model size in number of parameters for the different tasks (exact numbers can vary slightly due to variable vocabulary sizes with different data sets. G2P, MI and TR numbers correspond to the "large" configuration in \citet{wu-etal-2018-attention}.}
            \label{tab:params}
        \end{table}

\end{document}